\documentclass[letterpaper, 10 pt, conference]{ieeeconf}  % Comment this line out if you need a4paper

\IEEEoverridecommandlockouts                              % This command is only needed if 
\usepackage{graphics} % for pdf, bitmapped graphics files
\usepackage{subcaption}
\usepackage[british]{babel}
\usepackage{multirow}
\usepackage[hyphens]{url}
\usepackage{hyperref}
\usepackage{cleveref}
\usepackage{cite}
\usepackage{epsfig} % for postscript graphics files
\usepackage{mathptmx} % assumes new font selection scheme installed
\usepackage{times} % assumes new font selection scheme installed
\usepackage{amsmath,amssymb,amsfonts,mathrsfs} % assumes amsmath package installed
\usepackage{txfonts} % use varmathbb
\usepackage{multirow}

\usepackage{todonotes}
\usepackage{soul}
\definecolor{smoothgreen}{rgb}{0.7,1,0.7}
\sethlcolor{smoothgreen}

\title{\LARGE \bf
An Empirical Study of Person Re-Identification with Attributes
}

\author{Vikram Shree$^{1}$, Wei-Lun Chao$^{2}$ and Mark Campbell$^{1}$% <-this % stops a space
\thanks{$^{1}$Vikram Shree and Mark Campbell are with the Sibley School of Mechanical and Aerospace, Cornell University, USA
        {\tt\small \{vs476, mc288\}@cornell.edu}}%
\thanks{$^{2}$Wei-Lun Chao is with the Department of Computer Science, Cornell University, USA
        {\tt\small {wei-lun.chao}@cornell.edu}}%
}

\begin{document}

\maketitle
\thispagestyle{empty}
\pagestyle{empty}

%%%%%%%%%%%%%%%%%%%%%%%%%%%%%%%%%%%%%%%%%%%%%%%%%%%%%%%%%%%%%%%%%%%%%%%%%%%%%%%%
% \hidetodos
\begin{abstract}

Person re-identification aims to identify a person from an image collection, given one image of that person as the query.
There is, however, a plethora of real-life scenarios where we may not have a priori library of query images and therefore must rely on information from other modalities.
In this paper, an attribute-based approach is proposed where the person of interest (POI) is described by a set of visual attributes, which are used to perform the search.
We compare multiple algorithms and analyze how the quality of attributes impacts the performance. While prior work mostly relies on high precision attributes annotated by experts, we conduct a human-subject study and reveal that certain visual attributes could not be consistently described by human observers, making them less reliable in real applications. 
A key conclusion is that the performance achieved by non-expert attributes, instead of expert-annotated ones, is a more faithful indicator of the status quo of attribute-based approaches for person re-identification.

\end{abstract}

%%%%%%%%%%%%%%%%%%%%%%%%%%%%%%%%%%%%%%%%%%%%%%%%%%%%%%%%%%%%%%%%%%%%%%%%%%%%%%%%
\section{INTRODUCTION}
\label{sec:introduction}
%\todohere{Harry: Check the technical content and language!}

Accurate identification of people in crowded environments plays an indispensable role in various applications such as pedestrian tracking and surveillance. The specific task of person re-identification (re-ID) is to match images of people across diverse scenes, taken from different camera views or spatial locations.

One common assumption of person re-ID is the accessibility to a set of probe (query) images, which are to be matched to a different set of gallery images. Prior work has focused on extracting discriminative features from individuals' appearances \cite{matsukawa2016hierarchical, liao2015person, yang2014salient} and designing (or learning) appropriate distance metrics for feature matching \cite{liao2015person, xiong2014person, hirzer2012relaxed, roth2014mahalanobis, paisitkriangkrai2015learning}. More recently, deep convolutional neural networks (CNNs) have been widely applied for the re-ID problem because of their flexible architectures which jointly learn the two stages~\cite{wang2018resource, ahmed2015improved, yi2014deep, varior2016gated}, significantly improving the overall performance.

%The computer vision community has approached the problem by developing discriminative feature representations to encode the appearance of an individual \cite{matsukawa2016hierarchical, liao2015person, yang2014salient}. To further enhance the discriminative ability of extracted features, a high-dimensional distance metric is learnt from labelled data \cite{liao2015person, xiong2014person, hirzer2012relaxed, roth2014mahalanobis, paisitkriangkrai2015learning}. More recently, deep convolutional neural networks (CNNs) have been applied for re-ID because of their notable ability to capture semantic information \cite{wang2018resource, ahmed2015improved, yi2014deep, varior2016gated}. These methods combine feature extraction and metric learning into a single framework. As a consequence of success of deep learning in re-ID, it has been adopted by the robotics community for solving the well-known data association problem \cite{tang2017multiple}, \cite{zheng2016mars}.

\begin{figure}
    \centering
    \includegraphics[trim= 90 0 47 50, clip, width=.48\textwidth]{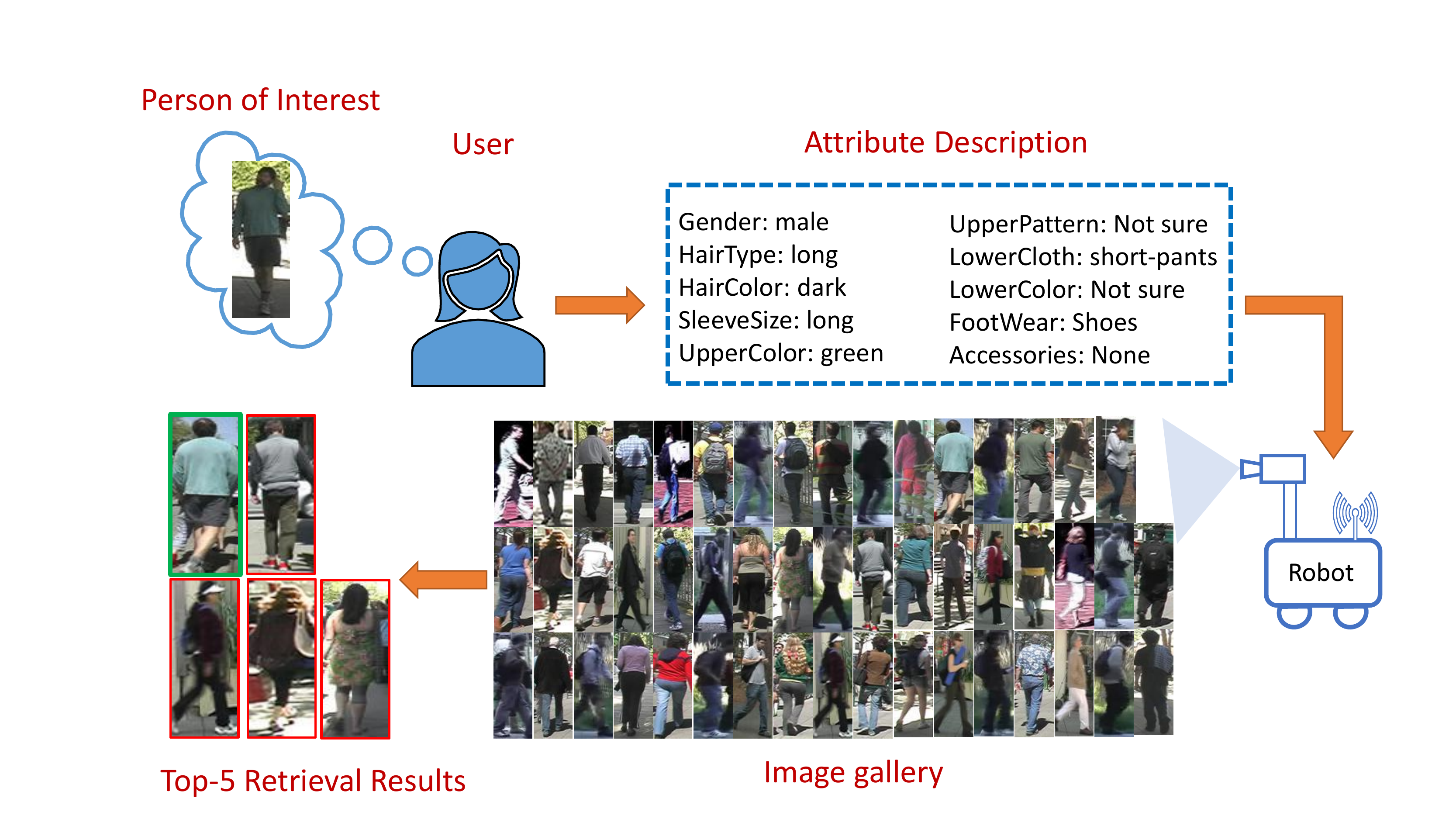}
    % \caption{\small Given a query image, the user describes the attributes of the person. The computer/robot retrieves the \textit{top-5} most relevant images from the gallery.}
    \caption{\small Illustration of person of interest (POI) search problem. A user describes the attributes that characterize the POI. A robot, by looking at the scene, creates a gallery of person-images. Based upon the attribute input, the robot returns the \textit{top-5} most relevant images from the gallery.}
    \label{fig:exapmleResult}
    \vspace{-0.2in}
\end{figure}

In contrast to typical re-ID problem, there are numerous scenarios in which there is no access to the query image of the person whom we want to identify. 
% This problem especially pops up in the areas of search and rescue, surveillance and suspect finding.
This problem is important in applications such as search and rescue, surveillance and suspect finding.
Without prior images, additional information is required to initialize the search.
It is hypothesized that visual attributes such as clothing, hair color, and footwear type embody rich semantic information about a person's visual appearance and could serve as appearance descriptions.
% One promising choice is the appearance descriptions provided by a human, who has seen the person of interest.
% As a result, we will have to rely upon high level appearance information provided by someone who has seen the person of interest.
% Since, visual attributes like clothing, hair-color, footwear-type, embody organized semantic information about a person's visual appearance, they could serve as appearance descriptions. 
% This problem setting is particularly relevant in the areas of search and rescue, and surveillance.
For example, consider a situation where person `A' is trapped in a building due to an unfortunate event and their friend  `B' wants to search them with a camera-mounted robot. 
To accomplish this task autonomously, the robot must rely upon the appearance information provided by person `B'; this assumes that an image of person `A' is not available at the moment. 
Even if an image is available, it may not match the current physical attributes of person `A'.
% Even if an image is available, the current physical attributes of person ``A", such as clothing, are mostly not known.
% Even if an image is accessible, it may not capture the current physical attributes of person ``A", such as clothing.
We abstract this task to be an attribute-to-image search problem where the robot searches for person `A' in the gallery of person-images from the scene, as illustrated in Figure \ref{fig:exapmleResult}. %\harry{This sentence is either redundant or misleading. Image retrieval is indeed image-to-image search.}
Since the task is no longer an image-to-image comparison, conventional re-ID methods become inadequate.

% Removing the query images introduces two main challenges into the re-ID problem. First, the query-data and the gallery images lie in different domains. This multi-modality in the data complicates the learning task. Second, since people have to rely upon their recollection of the attributes, it introduces uncertainty into the problem. 

Removing the query images introduces two main challenges to the re-ID problem. First, the query-data and the gallery images lie in different domains. This multi-modality in the data complicates the learning task. 
To address this problem, we leverage zero-shot learning (ZSL) methods which focus on associating data from different modalities \cite{lampert2014attribute, romera2015embarrassingly, changpinyo2017predicting, changpinyo2018classifier}; thus, the approach is referred to as \textit{zero-shot re-identification} \cite{roth2014exploration, layne2014attributes}.
The success of zero-shot re-ID depends highly upon the feature representation that we choose for the images. For this, we leverage a deep-CNN \cite{wang2018resource} which extracts highly discriminative features from the gallery images. %\harry{The second seems not an issue specifically for ZSL re-ID: it is common to general re-ID as well.}
% \vikram{The second problem is not a big concern in image-to-image re-ID because it does not involve people in it.}

% Second, since people have to rely upon their recollection of the attributes, it introduces uncertainty into the problem. 

The second challenge is uncertainty. It is said that, ``A picture is worth a thousand words", implying that there is inherent uncertainty when people are asked to describe the appearance of a person. The uncertainty becomes even more severe when people provide the visual attributes based upon their recollection.
% The uncertainty in data arises because certain attributes may not be distinctive enough for them to be observed and memorized by different people. 
Consequently, not all the attributes are correctly reported by the user during real-world applications and could hamper the performance of the zero-shot re-ID system. For example, in Figure \ref{fig:exapmleResult}, the user is not sure about the pattern of the clothing that the POI is wearing. A key limitation of the recent works in zero-shot re-ID is their reliance on expert-labelled attributes, obtained in a laboratory setting, which do not account for such uncertainties that may arise in the wild.
To address this issue, we present an exploratory human-subject study to select the most distinctive attributes for training and testing ZSL models. The major contributions of this paper can be summarized as follows:
\begin{itemize}
    \item First, in order to perform attribute-based person re-ID, we utilize a state-of-the-art feature representation for the images and evaluate the performance of several representative ZSL models on publicly available datasets.
    % \item First, we leverage a state-of-the-art feature representation for people-images and evaluate the performance of different ZSL models on publicly available datasets. Our results outperform the latest zero-shot re-ID approach.
    
    % Our results outperform the prior works in zero-shot re-ID.
%   \item First, in order to we have utilized a state-of-the-art feature representation for the images and evaluated several baseline ZSL models on publicly available dataset. 
  
%   \item First, we propose a novel architecture to perform zero-shot re-ID with human-interpretable attributes. We evaluate our method on publicly available datasets, based on labelled attributes and show that our approach improves the performance over the state-of-the-art method.
  
  \item Second, we design and conduct a human-subject study to identify key attributes that are consistently reported correct by various users. These are defined as \textit{distinct} attributes.

  \item Finally, we leverage the \textit{distinct} attributes for training ZSL models. The performance of zero-shot re-ID is evaluated using non-expert attributes and compared with that obtained from expert labels.
  
%   \item Finally, we leverage only the \textit{distinct} attributes for training the ZSL models and evaluate the performance with human-annotated attributes.

%   \item Finally, we train our regression model with the \textit{distinct} attributes and conducted a zero-shot re-ID experiment with human input data. Comparing the performance with results obtained from labelled attributes demonstrate that our model is robust to missing and incorrect attributes.
%   \item Second, we design and conduct a human-study to analyze their ability to memorize the attributes from images. \todohere{Change this line}
%   \item Finally, we utilize the most-memorizable attributes for training our regression model. To quantify the sensetivity of our model, we conduct the zero-shot re-ID experiment with human subjects and compare it with the results obtained based upon the expert's attributes.
\end{itemize}

\section{RELATED WORK}
\label{sec:relatedWork}
% \todohere{Harry: Check the technical content!}
% \todohere{Revise Section}
% We briefly review prior work on person re-ID and application of attributes for it.
% use of attributes for re-ID.

\subsection{Person Re-ID}
\label{subsec:personReid}
Traditional methods for re-ID have mainly focused on searching for better hand-crafted features such as chromatic content, spatial arrangement of colors, texture etc. \cite{farenzena2010person}, \cite{liao2015person}, \cite{hirzer2011person}. The intent is to find features that are mostly invariant to changes in pose, viewpoint and lighting conditions. Some hand-crafted features have the advantage of being human-interpreteble, for example visual attributes, 
and can serve as appearance description for zero-shot re-ID.
% and are suitable for the zero-shot re-ID problem. 
This is followed by a metric learning step which maps the features into a new space where feature vectors corresponding to same person are close to each other \cite{xiong2014person}, \cite{hirzer2012relaxed}, \cite{liao2015efficient}. 
% Unsupervised learning techniques are more scalable as compared to their supervised counterparts because they have less data dependence. 
Hand-crafted features, however, are usually not discriminative enough for differentiating idetities, leading to poorer performance.
% however, they yield poorer performance.

%\todohere{[HARRY]  The paragraph relating to CNN deals mainly with work [8] (2018), [24] (2017), [10] (2014) and [25] (2017). These works are represented as state of the art for deep representation learning and deep metric learning. However, there are more recent works in both areas that pursue better approaches. For example, https://arxiv.org/pdf/1804.01438.pdf and https://arxiv.org/abs/1704.01719 . (3) Later, in the same paragraph, the scalability of these methods with respect to the size of the required data sets is criticized. New methods like Triplet Hard Loss (https://arxiv.org/abs/1703.07737) or https://arxiv.org/abs/1809.05864 are conceived to perform well even on small training data sets (small number of person classes).}

Recently, supervised learning frameworks comprising of CNNs have been used for re-ID because of their ability to capture semantic and spatial information \cite{ahmed2015improved, cheng2016person, li2014deepreid, wang2018resource, zheng2017unlabeled}. Broadly, the methods can be divided into two categories: deep representation learning and deep metric learning. The first aims at creating a discriminative feature representation for the images \cite{xiao2016learning, wang2018resource, wang2018learning}. In \cite{xiao2016learning}, a robust feature embedding is learnt by training the model in multiple domains with domain guided dropout. 
In \cite{wang2018resource}, the authors fuse multiple embeddings across layers to capture spatial details lost in the last layer.
% In \cite{wang2018resource}, the authors propose that the last layer loses fine spatial details and fused multiple embeddings across layers to capture them. 
In contrast, deep metric learning intends to learn the similarity between images belonging to the same person \cite{yi2014deep, chen2017multi, chen2017beyond}. In \cite{yi2014deep}, a siamese CNN model is leveraged for jointly learning the features and the metric for re-ID. In \cite{chen2017multi}, a multi-task method is proposed to integrate the classification and ranking task together. 
The state-of-the-art supervised learning approaches for re-ID have achieved formidable feature representation capability. However, there is still skepticism about the scalability because of their heavy reliance on large amount of training data, demanding careful design of learning objectives and network architectures~\cite{zhai2019defense, hermans2017defense}. Besides, their inadequacy in handling multi-modal data paves the way for ZSL models in zero-shot re-ID problem.
% Although, the state-of-the-art supervised learning methods for re-ID have achieved formidable performance, there is still skepticism about scalability because of their heavy reliance on large amount of training data.

% Attributes review starts here
\subsection{Application of Attributes in Re-ID}
\label{subsec:attReId}

Visual attributes such as clothing, hair-style, shoe-type, accessories etc, have been used as mid-level feature representation for re-ID in a number of works \cite{matsukawa2016person, lin2017improving, schumann2017person, roth2014exploration, wang2018transferable}.  In \cite{lin2017improving}, the authors propose an attribute-recognition network which combines ID classification and attribute classification losses. Because the attributes are human-interpretable, they are useful for zero-shot re-ID \cite{layne2014attributes, roth2014exploration}. In \cite{layne2014attributes}, the authors train an attribute-classifier and use the attributes as feature representation. Further, a distance metric is used for comparing similarity of images in the attribute space. In \cite{roth2014exploration}, a clustering scheme is used to determine the useful attributes in order to maximize re-ID performance. Instead of just relying upon attributes, a few authors utilize them as auxiliary information, thus inhibiting their application to the zero-shot re-ID setting. In \cite{matsukawa2016person}, features from the CNN are fine-tuned on a pedestrian attribute dataset by adding a combination attribute loss term.

% However, the attribute-centric approach leads to inferior performance as compared to deep-features. This is mainly because high-quality attribute prediction is difficult when training data is sparse, and the images have low resolution. This is our main motivation behind abstaining from using attributes as the final feature representation for person-images and rather taking the representation route. 

% In this work, we leverage attributes for approaching the zero-shot re-ID problem. 
In this work, we leverage visual attributes for characterizing person-appearance in the zero-shot re-ID problem.
However, methods relying on attribute detection achieve inferior performance because high-quality attribute prediction is difficult when training data is sparse and images have low resolution. This is our primary motivation for taking a representation-learning route, where the gallery images are projected into a discriminative feature embedding, followed by a ZSL model which associates the query attributes to the projected gallery features.
% second, additional step of metric learning involved.

% However, attribute-centric feature representation for re-ID leads to inferior performance as compared to conventional feature vector. This is mainly because high-quality attribute prediction is difficult when training data is sparse and the images have low resolution. This is our main motivation behind transforming the attribute-features into the deep feature space. 

% \todohere{Define problem definition.}

\section{PROBLEM FORMULATION}
\label{sec:formulation}

% Consider a gallery with $N$ images, with $K$ distinct person classes within it. Let us assume that the query set consists of attributes for $M$ people to be searched in the gallery. Define $\mathcal{X}_{f}^{g} = \{\mathbf{z}_{f}^{i} | i = 1, \hdots, N \}$, as the set of features for the gallery images and $\mathbf{U}^{g}$ denotes the associated identity labels i.e. $\mathbf{U}^{g} = [u^{1},\hdots,u^{N}]^{T}$, where $u^{i} \in \{1, \hdots, K\}$. 
% Similarly, define $\mathcal{X}_{A}^{q} = \{\mathbf{x}_{A}^{j} | j = 1, \hdots, M \}$, as the set of attributes for query people, each being $d_{a} \times 1$ dimensional. 
% % Assuming that all queries belong to a person-class in the gallery, let $\mathbf{V}^{q}$ represent the vector of labels for the queries i.e. $\mathbf{V}^{q} = [v^{1},\hdots,v^{M}]^{T}$, where $v^{j} \in \{1, \hdots, K\}$. 

% A ZSL module learns a classifier function $\mathbf{f}_{k}: \varmathbb{R}^{d_a} \rightarrow \varmathbb{R}$, which yields the likelihood of the query attributes belonging to each person-class $k$. This is followed by predicting labels $\hat{v}^{j}$ for the queries, as follows:

% \begin{equation*}
%     \hat{v}^{j} = \arg\max_{k \in \{1, \hdots, K \}} \mathbf{f}_{k}(\mathbf{x}_{A}^{j})
%     % \boldsymbol{\eta}_p=\mathbf{f}(\mathbf{z}^\text{app}_p),\,\forall p =1,\dots,N.
% \end{equation*}

% Vikram rewrote this section
Consider a gallery with $K$ distinct person images within it. Let us assume that the query set consists of attributes for $M$ people to be searched in the gallery. Define $\mathcal{X}_{f}^{g} = \{\mathbf{z}_{f}^{i} | i = 1, \hdots, K \}$, as the set of features for the gallery images, each being $d_{f} \times 1$ dimensional and $\mathbf{U}^{g}$ denotes the associated identity labels i.e. $\mathbf{U}^{g} = [u^{1},\hdots,u^{N}]^{T}$, where $u^{i} \in \{1, \hdots, K\}$. 
Similarly, define $\mathcal{X}_{A}^{q} = \{\mathbf{x}_{A}^{j} | j = 1, \hdots, M \}$, as the set of attributes for query people, each being $d_{a} \times 1$ dimensional. 
% Assuming that all queries belong to a person-class in the gallery, let $\mathbf{V}^{q}$ represent the vector of labels for the queries i.e. $\mathbf{V}^{q} = [v^{1},\hdots,v^{M}]^{T}$, where $v^{j} \in \{1, \hdots, K\}$. 

A ZSL module learns a classifier function $\mathbf{f}_{k}: (\varmathbb{R}^{d_a} \times \varmathbb{R}^{d_f} \times \hdots \times \varmathbb{R}^{d_f}) \rightarrow \varmathbb{R}$, which yields the likelihood of the query attributes belonging to each person-class $k$. This is followed by predicting labels $\hat{v}^{j}$ for the queries, as follows:

\begin{equation*}
    \hat{v}^{j} = \arg\max_{k \in \{1, \hdots, K \}} \mathbf{f}_{k}(\mathbf{x}_{A}^{j}, \mathbf{z}_{f}^{1}, \hdots, \mathbf{z}_{f}^{K})
    % \boldsymbol{\eta}_p=\mathbf{f}(\mathbf{z}^\text{app}_p),\,\forall p =1,\dots,N.
\end{equation*}

\section{PEDESTRIAN ATTRIBUTES}
\label{sec:pedAttributes}
% \todohere{Revise Section}
% \textbf{$<$Explain which features are we using and how are we representing the features in attribute space. Show the table for attributes.$>$}

Visual attributes encode high level appearance information which enables humans to distinguish between different people e.g. clothing, shoes, hair-style etc. Since hand-labelling attributes is a time-consuming and tedious process, in this work, we leverage the attribute labels provided in Pedestrian Attribute dataset (PETA) \cite{deng2014pedestrian}.
PETA is a large public collection of images with labelled attributes; consisting of 61 binary and 4 multi-class attribute labels. 
% Since hand-labelling images is a time-consuming and tedious process, in this work, we leverage the attributes provided by PETA.

We divide the attributes in PETA into 13 mutually exclusive classes, each one consisting of its own list of attributes. It is known that a high dimensional function is more difficult to learn than a low dimensional function in the absence of sufficient data. In order to simplify the learning problem, we prune the attributes. In particular, we remove rare attributes in the dataset and categorize multiple basic attributes under a common meta-attribute e.g. top clothing colors like black, brown, and purple have been combined into a single attribute called `UpperBodyDark'. Furthermore, we exclude certain attributes e.g. age, due to the difficulty in its consistent prediction based upon the low-resolution images in the dataset. The final set of attributes and meta-attributes in each class is presented in Table \ref{tbl:attributes}. The attributes that constitute the meta-attributes are listed in Table \ref{tbl:metaAttributes}.

% Based upon the presence or absence of an attribute or meta-attribute in an image, we create a binary feature vector in the attribute space. As a result, we get a 44 dimensional vector in the attribute space, comprising of `0' and `1'.

In this work, all attributes and meta-attributes belonging to a class are encoded by binary values. As a result, a 44 dimensional vector $\mathbf{x}_{A}^{j}$ is obtained in the attribute space, comprising of `0' and `1', where a `0' signifies the absence of an attribute/meta-attribute in the image and `1' signifies its presence in the image. It should be noted that considering continuous attribute values could be useful for certain classes, like UpperColor; however, we were limited by the binary nature of the labelled data in PETA.
% \todohere{The above paragraph should not be a part of this section.}

% \begin{table}[h]
% \centering
% \caption{Pedestrian Attributes categorized into Mutually Exclusive Classes}
% \begin{tabular}{|l|p{4.5cm}|p{1.4cm}|}
% \hline
%  \textbf{Classes} & \textbf{Attributes or Meta-attributes} & \textbf{Number of Attributes} \\ \hline
%  Gender & female, male  & 2 \\ \hline
%  HairType & short, long, bald  & 3 \\ \hline
%  HairColor & \textit{binaryDark}, \textit{binaryLight}  & 2 \\ \hline
%  UpperCloth & jacket, suit, sweater, t-shirt  & 4 \\ \hline
%  SleeveSize & long, short, noSleeve  & 3 \\ \hline
%  UpperColor & red, blue, green, \textit{dark}, \textit{light}  & 5 \\ \hline
%  UpperPattern & \textit{stripes}, plaid, logo  & 3 \\ \hline
%  LowerCloth & formalPants, \textit{informalPants}, \textit{shortPants}, skirt  & 4 \\ \hline
%  LowerColor & red, blue, green, \textit{dark}, \textit{light}  & 5 \\ \hline
%  FootwearType & sandals, \textit{allShoes}, boots & 3 \\ \hline
%  FootwearColor & binaryDark, binaryLight & 2 \\ \hline
%  Carrying & backpack, messengerBag, plasticBag, folder & 4 \\ \hline
%  Accessory & headphones, sunglasses, hairband, hat & 4 \\ \hline
%  \multicolumn{2}{|c|} {\bfseries Total number of attributes} & \textbf{44}\\ \hline
 
% \end{tabular}
% \label{tbl:attributes}
% \end{table}

\begin{table}
\vspace{0.1in}
\centering
\caption{\small Pedestrian attributes categorized into Mutually Exclusive Classes.}
% \begin{small}
\begin{tabular}{|l|l|p{3.4cm}|p{1.3cm}|}
\hline
~ & \textbf{Classes} & \textbf{Attributes or Meta-attributes} & \textbf{Number of Attributes} \\ \hline
 Q1 & Gender & female, male  & 2 \\ \hline
 Q2 & HairType & short, long, bald  & 3 \\ \hline
 Q3 & HairColor & \textit{binaryDark}, \textit{binaryLight}  & 2 \\ \hline
 Q4 & UpperCloth & jacket, suit, sweater, t-shirt  & 4 \\ \hline
 Q5 & SleeveSize & long, short, noSleeve  & 3 \\ \hline
 Q6 & UpperColor & red, blue, green, \textit{dark}, \textit{light}  & 5 \\ \hline
 Q7 & UpperPattern & \textit{stripes}, plaid, logo  & 3 \\ \hline
 Q8 & LowerCloth & formalPants, \textit{informalPants}, \textit{shortPants}, skirt  & 4 \\ \hline
 Q9 & LowerColor & red, blue, green, \textit{dark}, \textit{light}  & 5 \\ \hline
 Q10 & FootwearType & sandals, \textit{allShoes}, boots & 3 \\ \hline
 Q11 & FootwearColor & binaryDark, binaryLight & 2 \\ \hline
 Q12 & Carrying & backpack, messengerBag, plasticBag, folder & 4 \\ \hline
 Q13 & Accessory & headphones, sunglasses, hairband, hat & 4 \\ \hline
 \multicolumn{3}{|c|} {\bfseries Total number of attributes} & \textbf{44}\\ \hline
\end{tabular}
\label{tbl:attributes}
% \vspace{-0.1in}
% \end{small} 
\end{table}

\begin{table}
\centering
\caption{\small Collection of basic attributes combined to create meta-attributes.}
\begin{tabular}{|l|l|}
\hline
 \textbf{Meta-attributes} & \textbf{Basic attributes}\\ \hline
 \textit{binaryDark} & black, blue, purple, brown \\ \hline
 \textit{binaryLight} & white, pink, orange, green, yellow, grey, red \\ \hline
 \textit{dark} & black, purple, brown \\ \hline
 \textit{light} & white, pink, orange, yellow, grey \\ \hline
 \textit{informalPants} & jeans, trousers \\ \hline
 \textit{shortPants} & shorts, capri, hotPants \\ \hline
 \textit{allShoes} & shoes, sneakers, leatherShoes \\ \hline
 \textit{stripes} & thickStripes, thinStripes\\ \hline
 
\end{tabular}
\label{tbl:metaAttributes}
\vspace{-0.2in}
\end{table}

% \section{ZERO-SHOT RE-ID WITH ATTRIBUTES}
\section{APPROACH}
\label{sec:zeroshotReid}
% \todohere{Harry: Check the technical content!}

% \todohere{Check this section with Harry.}

\begin{figure}
\vspace{0.1in}
    \centering
    \includegraphics[trim= 15 90 270 75, clip, width=.48\textwidth]{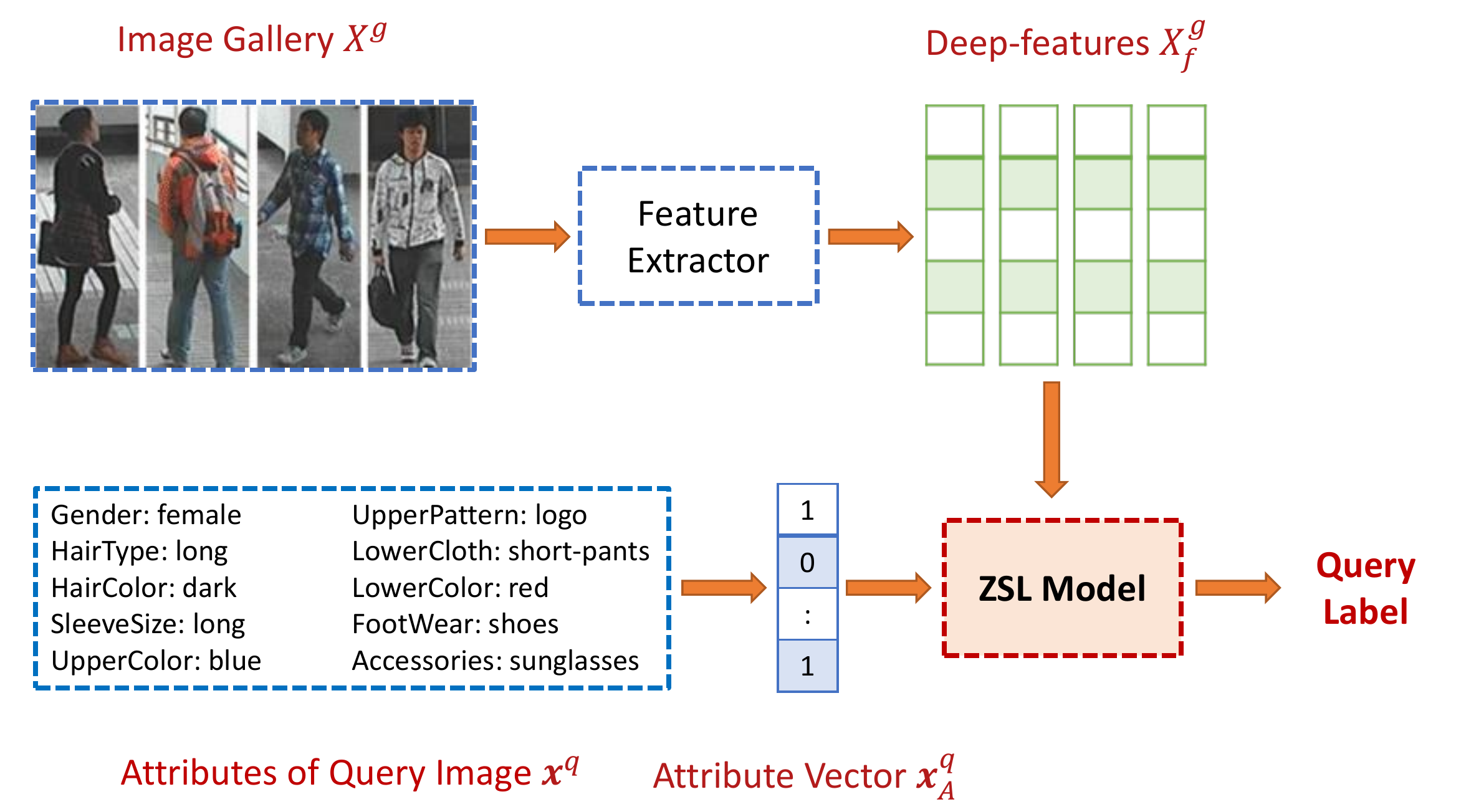}
    % \caption{\small Given a query image, the user describes the attributes of the person. The computer/robot retrieves the \textit{top-5} most relevant images from the gallery.}
    \caption{\small Illustration of a general zero-shot re-ID pipeline. The feature extractor maps the gallery images into semantic embedding and ZSL models learns to associate the attributes with the predicted semantic embedding, thus assigning labels for the query attributes. }
    % \todohere{change the Xdg and XAq in the figure.}}
    \label{fig:zslPipeline}
    \vspace{-0.2in}
\end{figure}

We describe the framework for addressing the zero-shot re-ID, where the task is to classify the query attributes into label space of the gallery images. A general approach is shown in Figure \ref{fig:zslPipeline}. There are two important steps involved in zero-shot re-ID. The first step is to learn a discriminative feature representation for the people-images, referred to as the feature extractor. The second step involves learning to associate the attributes with the deep-feature representations. In this section, we describe both the steps, while assuming that labelled attributes are available for the query images.   

\subsection{Learning Feature Representation}
\label{subsec:dare}

The goal of feature learning is to transform the images into semantic embedding space where the Euclidean distance between features belonging to the same person is smaller than those of different people. As mentioned in Section \ref{sec:relatedWork}, several deep-CNN models have been proposed in order to learn feature representations for re-ID, from images. In this work, we leverage Deep Anytime Re-ID (DaRe) \cite{wang2018resource}, a state-of-the-art re-ID framework which also allows trade-off between computational resource requirement versus performance; this makes it suitable for robotics applications. The DaRe architecture consists of several sequential, convolutional stages and finally, a fusion stage. Information is fused across multiple layers to capture both coarse semantic information and fine-level details. To achieve this, the feature maps at intermediate layers are first passed through a fully connected layer, which brings them to the same dimension. For a given image $\mathbf{x}$, assume that $\mathbf{\phi}_{s}(\mathbf{x})$ denotes the embedding produced at stage $s$. Subsequently, the deep-feature $\mathcal{X}_{d}$ is obtained by doing a weighted sum of the intermediate embeddings:

\begin{equation}
    \mathcal{X}_{d} = \mathbf{\phi}_{fusion}(\mathbf{x}) =
    \sum_{s=1}^{S} w_{s} \mathbf{\phi}_{s}(\mathbf{x}),
\end{equation}
where $S$ is the total number of stages in the network and $w_{s}$ are learnable parameters. The overall loss function used for training the network consists of individual loss functions for each of the layers and a final fusion loss function. Each loss function is computed based on triplet loss since \cite{hermans2017defense} showed that it has superior performance as compared to conventional surrogate losses for the re-ID task. 

% \begin{equation}
%     {l}_{total} =
%     \sum_{s=1}^{S} l_{s} + l_{fusion}
%     \label{eq:loss}
% \end{equation}

% subsection{Attribute to feature association}

% Motivated by \cite{hermans2017defense}, we use the triplet loss for each loss function in equation \ref{eq:loss}. 
% Triplet loss is used for each loss function in Eq.~(\ref{eq:loss}), since \cite{hermans2017defense} showed that it has superior performance as compared to conventional surrogate losses for the re-ID task. 
% \harry{This sentence might be misleading. I think \cite{wang2018resource} already used triplet loss.} 
% The triplet loss function encourages the farthest positive examples to be closer and forces the closest negative examples to be farther apart in the feature space. Please be referred to \cite{wang2018resource} for more details. 
% \harry{I won't suggest going to this technical detail of DaRe. It's better to give readers a concrete and concise idea. In the current version, the $l_s$ and $l_{fusion}$ are not well-defined, and more importantly, the loss is computed on conventional re-ID, which might be confusing with the paper's goal.}

\subsection{Associating Attributes with Deep-Features}
\label{subsec:ZSLapproaches}

Numerous techniques have been proposed in ZSL literature to associate the query-attributes to the visual features of the gallery~\cite{xian2018zero,changpinyo2018classifier}. We describe three broad classes of ZSL algorithms, different by how the association is achieved.

%\todohere{[HARRY] The four chosen approaches evaluated should be discussed in greater detail. It remains unclear if the authors used indeed exact implementations of the approaches. They use phrases like "For example, DAP [12]..." making it unclear if indeed they used DAP "as is" or present an implementation only loosely inspired by said work. The authors should furthermore justify why they included these four methods specifically.}

\subsubsection{Learning to predict visual embedding}
Given the query-attributes $\mathcal{X}_{A}^{q}$, one can learn to predict the visual embedding i.e. the deep-feature vectors $\mathcal{X}_{d}^{q}$. Consequently, the identity-label of the query can be obtained using a similarity measure between the predicted embedding $\mathcal{X}_{d}^{q}$ and the gallery-features $\mathcal{X}_{d}^{g}$. In this paper we apply EXEM \cite{changpinyo2018classifier}, a state-of-the-art ZSL algorithm that uses support vector regression to predict visual embedding, followed by nearest neighbor in the Euclidean space for label assignment.

\subsubsection{Learning to predict attributes}
Given the deep-feature representation of the gallery images $\mathcal{X}_{d}^{g}$, one can learn to predict the attributes $\mathcal{X}_{A}^{g}$. Based upon the similarity between the predicted attributes $\mathcal{X}_{A}^{g}$ for gallery images and the given query attributes $\mathcal{X}_{A}^{q}$, the query-label can be estimated. In this paper we apply DAP \cite{lampert2014attribute}, which learns a probabilistic classifier to predict each attribute and labels the query by MAP assignment. The main difference between EXEM and DAP is the space in which the similarity is computed: EXEM computes similarity in the visual space, while DAP does so in the attribute space.

\subsubsection{Compatibility learning}

In this approach, a common representation is learned, onto which both the query-attributes and deep-features of the gallery, are projected. This is followed by an optimization which maximizes the compatibility score of the projected instances in the space. Different methods under this category differ in their choice of the common space or the compatibility function. In this paper we apply ESZSL \cite{romera2015embarrassingly}, which uses bi-linear mapping for the projection and has an efficient closed form solution.

\section{EXPERIMENTAL EVALUATION}
\subsection{Dataset and Implementation Details}
\label{subsec:reidDataset}
% \WLC{Put this as a separate section.}
% \textbf{$<$Explain the dataset that is used and its division into train, validation and test splits. Explain the training process of DARE network. Compare results with baseline.$>$}

% We evaluate the different zero-shot re-ID approaches on two publicly available datasets: a) VIPeR \cite{gray2007evaluating} and b) PRID \cite{hirzer2011person}.

We implement three ZSL algorithms---EXEM~\cite{changpinyo2018classifier}, DAP~\cite{lampert2014attribute}, and ESZSL~\cite{romera2015embarrassingly}---as described in Section~\ref{subsec:ZSLapproaches}, and evaluate the performance with labelled attributes on two public datasets: a) VIPeR \cite{gray2007evaluating} and b) PRID \cite{hirzer2011person}. 

% We evaluate a representative algorithm belonging to each category of ZSL methods, as mentioned in \ref{subsec:ZSLapproaches}, on two publicly available datasets: a) VIPeR \cite{gray2007evaluating} and b) PRID \cite{hirzer2011person}.
% with the improved DaRe feature representation.

% We have evaluated our zero-shot re-ID approach, as well as the baselines on two publicly available datasets: a) VIPeR \cite{gray2007evaluating} and b) PRID \cite{hirzer2011person}.

The VIPeR dataset consists of 632 pedestrian image pairs taken from two cameras with significant variation in viewpoint and illumination. All the images in the dataset are scaled to 128$\times$48 pixels. The dataset is divided randomly into three sets: train, validation and test sets. There are 280 pedestrian image pairs in the train set, 36 in the validation set and 316 in the test set.

% \cite{layne2014attributes}
The PRID dataset consists of both single-shot and multi-shot images, taken from two cameras. Similar to \cite{layne2014attributes}, we only use the first 200 person-images for each camera, since they appear in both the cameras. All the images in the dataset are scaled to 128$\times$64 pixels. The dataset is divided randomly into three sets: train (80), validation (20) and test (100) sets.

% The attributes for VIPeR and PRID are provided in PETA dataset.

Following Section \ref{sec:pedAttributes}, $44$ dimensional attribute vectors are created for the images in VIPeR and PRID, based on expert-labels available from PETA dataset.
% All the ZSL models are trained with the $44$ attributes, as shown in Table \ref{tbl:attributes}, and the deep-features produced by DaRe network. 
The ZSL models are trained with the computed attribute vectors and the deep-features produced by DaRe network.
We utilize the ResNet-50 implementation of the DaRe for learning the deep-feature representation, training it with the same protocol as in \cite{wang2018resource}. 
\subsection{Results}
\label{subsec:reidPerformance}

We compare the performance of different zero-shot re-ID methods, obtained with the DaRe feature representation, to the previous state-of-the-art approach by \textit{Layne et al.} \cite{layne2014attributes}. The standard Rank-N \textit{Cumulative Matching Characteristic} accuracy is used as the performance metric.

% Table \ref{tbl:zeroshotPerformance} summarizes our results on both the datasets. For VIPeR dataset, we observe that the accuracy of all the approaches outperform the state-of-the-art performance by a significant margin. For PRID dataset, the KRR model is competitive to the state-of-the-art result, however, the other ZSL models mostly outperform \cite{layne2014attributes}. 

Table \ref{tbl:zeroshotPerformance} summarizes our results on both the datasets. For VIPeR dataset, we observe that all the representative ZSL approaches, DAP, ESZSL and EXEM, outperform the baseline by a significant margin. For PRID dataset, the ZSL models achieve superior accuracy at low rank values. However, at higher ranks, the accuracies are competitive.

% Table \ref{tbl:zeroshotPerformance} summarizes our results on both the datasets. For VIPeR dataset, we observe that all the ZSL baselines, including our KRR model outperform the state-of-the-art performance by a significant margin. For PRID dataset, our KRR model is competitive to the state-of-the-art result, however, the other ZSL models mostly outperform \cite{layne2014attributes}. 

% The improvement in performance can be attributes to the better deep-feature representation that DaRe enables us to establish.

The improvement over the state-of-the-art approach can be attributed to two main factors. First, DaRe features are more identity-discriminating than the previously used feature representations. Second, in contrast to other re-ID techniques which rely on very high dimensional features, DaRe features are fairly low dimensional (128). With scarce attribute data, learning a lower dimensional mapping from the attribute space to the deep-feature space is easier than learning a high dimensional mapping. Comparing among the three ZSL algorithms, we found that EXEM and ESZSL outperform DAP in general, which aligns well with the observations in object recognition~\cite{xian2018zero,changpinyo2018classifier}.

\begin{table}
\vspace{0.1in}
\centering
\caption{ \small \textit{Cumulative Matching Characteristic} accuracy for zero-shot re-ID on VIPeR and PRID dataset. Superior results are shown in \textbf{bold}. }
% \todohere{Search for better hyperparameters for KRR to evaluate on PRID. Train on validation set.}
\begin{tabular}{|l|l|l|l|l|l|l|}
\hline
~ & \textbf{Method} & \textbf{Rank-1} & \textbf{Rank-5} & \textbf{Rank-10} & \textbf{Rank-25} \\ \hline
\multirow{5}{*}{\rotatebox[origin=c]{90}{VIPeR}} & \textit{Layne et al.} \cite{layne2014attributes} & 6.0 & 17.1 & 26.0 & 48.1 \\ 
& DAP \cite{lampert2014attribute} & 7.0 & 25.6 & 34.5 & 56.7 \\
& ESZSL \cite{romera2015embarrassingly} & \textbf{8.9} & 27.2 & 41.5 & 61.1 \\ 
& EXEM \cite{changpinyo2018classifier} & 7.9 & \textbf{31.3} & \textbf{43.0} & \textbf{62.0} \\
% & KRR & 8.5 & 20.9 & 30.4 & 49.7 \\ 
\hline
\multirow{5}{*}{\rotatebox[origin=c]{90}{PRID}} & \textit{Layne et al.} \cite{layne2014attributes} & 8.0 & 29.0 & 47.0 & 73.0 \\ 
& DAP \cite{lampert2014attribute} & 12.0 & 38.0 & 50.0 & \textbf{78.0}\\
& ESZSL \cite{romera2015embarrassingly} & \textbf{14.0} & \textbf{48.0} & \textbf{55.0} & 76.0\\ 
& EXEM \cite{changpinyo2018classifier} & 12.0 & 32.0 & 47.0 & 72.0 \\
% & KRR & 9.0 & 32.0 & 45.0 & 72.0 \\ 
\hline
\end{tabular}
\label{tbl:zeroshotPerformance}
\vspace{-0.1in}
\end{table}

% \colorbox{orange!30}{\textcolor{red}{No more sub-sections in zero-shot Re-ID}}

% \subsection{Attribute-based Feature Representation}
% \subsection{Feature Representation}
% \subsection{Attributes to Feature Transformation}

% \section{Attribute Study}
% \section{Attribute Study to Identify Distinct Attributes}

\section{ATTRIBUTE SIGNIFICANCE STUDY}
\label{sec:attributeStudy}
% \todohere{Revise Section}

Previous works in attribute-based re-ID have relied on expert attributes. This is a valid assumption in a laboratory setting; however, in a real-world application, the user may not remember all the attributes corresponding to a person of interest. Different techniques have been proposed to handle missing attributes like mean-imputation \cite{donders2006gentle}; yet, having more imputed attributes can significantly impede re-ID performance. Furthermore, some of the attributes can be ambiguous and different people might perceive it differently e.g. a messenger-bag can be confused with a plastic-bag.

% Moreover, even if people remember certain attributes, there can be
% There are ways to deal with missing attributes e.g.
% These missing attributes can create ambiguity in the re-ID process and impede performance. 

In this work, rather than using all the attributes (Table \ref{tbl:attributes}), available in the PETA dataset, we propose to rely only upon \textit{distinct} attributes. The major drawback of using all the attributes is that ambiguities in attribute-inference during test will impede the performance of the trained re-ID module.
Consequently, we have defined \textit{distinct} attributes as the ones whose predictions consistently match the expert-labels.
% We define \textit{distinct} attributes as those that can be consistently inferred by different users during an actual deployment of our system in a realistic setting. 
This motivated us to perform a human-subject study that analyzes the significance of attributes from a non-expert's perspective. In this section, our experimental setup and data processing steps involved in the study are described.

% Rather than using all the attributes present in the dataset, we propose to use only the ``distinct" ones that are easy to remember for humans. 

\subsection{Research Question}
\label{subsec:hypo1}

% The central hypothesis of our study is that humans can perceive and recall certain visual attributes of people that they have encountered in the past.
% In this study we are testing 
The goal of this study is to evaluate whether humans can perceive and recall visual attributes of people, whose images they have seen before.

\textit{Hypothesis \textendash} 
% Given an image of a person, humans can identify and memorize the \textit{distinct} attributes associated with him/her, better than a randomized algorithm.
Given an image of a person, humans are better than a randomized algorithm at identifying and memorizing the \textit{distinct} visual attributes of the POI.

% Humans are better than a randomized algorithm at predicting \textit{distinct} attributes associated with people, whose images are

% The central hypothesis of our study is that humans can perceive and recall certain visual attributes associated to people, even if the actual person or their image is absent.
% We formalize the problem in a one-tailed \textit{t-}test framework with the following hypotheses:
% statistical testing framework 
% \begin{itemize}
%   \item \textbf{H0} (\textit{Null} hypothesis): Human's perception of visual attributes is no better than random guess.
%   \item \textbf{H1} (\textit{Alternate} hypothesis): Human's perception of visual features are better than random guess.
% \end{itemize}

\subsection{Experiment Design}
\label{subsec:survey1}
In order to test our hypothesis, we modelled a setup where participants were asked to look at an image of a person for a limited amount of time.
% In order to test our hypothesis, we conducted a survey with 10 participants where they were asked to look at an image of a person 
% % $I_{j}^{k}$, where $k$ denotes the participant number and $j$ denotes image number),
% for limited amount of time. 
This was followed by a questionnaire\footnote[3]{A sample of the questionnaire that was used in the study can be viewed at \href{https://github.com/vikshree/humanRobotReid}{\texttt{https://github.com/vikshree/humanRobotReid}}} about the attributes that are present in the image.
This setup is an attempt to model a real-world situation where a person is trying to find someone whom them have seen in the immediate past.
The limited time setting is important to identify the attributes that stand out to humans. In contrast, letting humans examine an image for extended period of time would allow them to memorize every detail in the image and thus defeat the purpose of the study.

% In contrast, letting them look into the image forever will allow them to memorize every detail in the image, especially in a survey setting.

Each participant repeated the task five times with different images, each taken from a different re-ID dataset i.e. VIPeR \cite{gray2007evaluating}, CUHK \cite{li2013locally}, PRID \cite{hirzer2011person}, TownCentre and iLIDS \cite{wang2014person}. 
% A total examination time of $15$ seconds is alloted for each image.
The examination time for each image was decided based on trial run of the study on 6 participants, where $4$ out of $6$ participants reported that $15$ seconds is enough time for them to identify and remember the attributes. Consequently, we fixed the examination time to be $15$ seconds for the actual study.
% and was allotted 15 seconds to look at each image. 
% The \textit{t}-test assumes that data samples are independent which is not strictly valid if the same participant repeats the task. However, since the images are different so, the independence assumption still holds to some extent. 
% There was no time constrain for answering the questionnaire.

The questionnaire consists of 13 questions, one corresponding to each attribute class listed in Table \ref{tbl:attributes}; the options represent the attributes present in that class. 
There was no time constraint for answering the questionnaire.
Not all classes used in the study have exhaustive set of attributes e.g. a person might wear a cloth type that is not present in the class `UpperBody'. As a result, we have included `None of the above' option for such classes. Broadly, the questions can be divided into two types: a) multiple correct type which have more than one correct choice e.g. accessory, and b) single correct type which have only one correct option e.g. gender. However, irrespective of the type of question, participants can always choose `Not sure' option, if they are doubtful. 

% \todohere{add github link to additional material}

%  Furthermore, every question has the option ``Not sure" and should be selected if the participant is doubtful about the choices

% \subsection{Performance Metric}
\subsection{Data Processing}
\label{subsec:dataProcessing}

To evaluate the performance of a participant in the study, we compare their response with the label attributes, available from PETA dataset, and assign scores based on the comparison. The following intuitive scoring scheme is used:

\begin{itemize}
  \item For single correct type questions, we assign 100 points for choosing the correct option and 0 points for choosing an incorrect option.
%   \item For single correct type questions, the participant gets 100 points for choosing the correct option and 0 points for choosing an incorrect option.
  \item For multiple correct type questions, we assign $\frac{100}{n}$ points for choosing each correct option and 0 points for the incorrect ones, where `$n$' is the number of correct options in the question.
  
%   \item For multiple correct type questions, the scoring is slightly different. Let us say that there are $n$ attributes present in the labelled data. For each correct attribute choice, we grant $\frac{100}{n}$ points to the participant and 0 points for each incorrect attribute choice. Thus, similar to single correct type questions, the maximum score is still upper bounded by 100.
\end{itemize}

% The limited time setup is important to understand the recall capability of different people. In contrast,

% Each participant repeated this task five times with different images (let's say $I_{1}^{k}, I_{2}^{k}, I_{3}^{k}, I_{4}^{k}, I_{5}^{k}$) and was allotted 10 seconds to look at each image.

% However, there was no time constraint for the questionnaire.

% In order to test our hypothesis, we conducted a survey with 10 participants where they were asked to look at an image of a person (let's say $I_{i}^{k}$, where $k$ denotes the participant number and $i$ denotes image number) for limited amount of time (10 seconds). This was followed by a questionnaire about the attributes that are present in $I_{i}^{k}$. 

\subsection{Results}
\label{subsec:survey1Results}
\begin{figure}
\vspace{0.1in}
    \centering
    \includegraphics[trim= 35 5 60 34, clip, width=.45\textwidth]{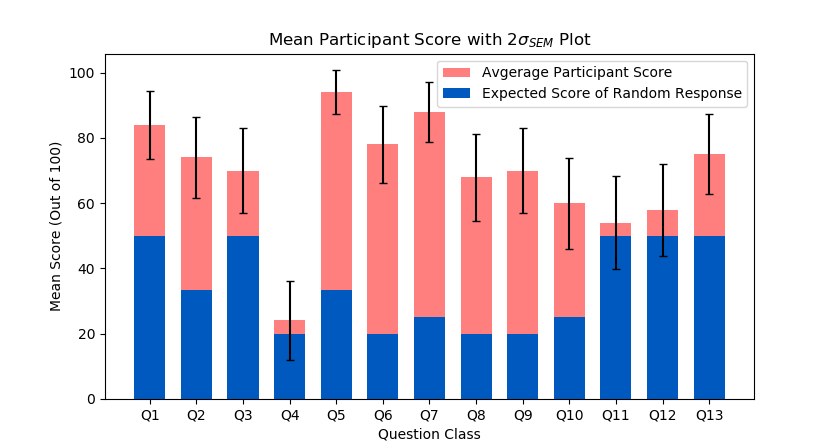}
    \caption{\textit{Mean} score of participants and \textit{expected} score of random guess for each attribute class. Error bars indicate 95$\%$ confidence intervals.}
    \label{fig:surveyBarPlot}
    % \vspace{-0.1in}
\end{figure}

The participation for the study was voluntary and a total of 10 participants completed it. All of them are graduate students, enrolled at Cornell University and belong to the age group of 20-30 years. Each participant examined and labelled five different images, sequentially. 
% Each one of them responded to questions belonging to 5 different images, we collected a total of 50 data samples. 
Figure \ref{fig:surveyBarPlot} shows the mean score of participants and expected score of a randomized algorithm.
We observe that the error bars for classes Q4, Q11 and Q12 overlap with the expected random score, indicating that humans perform poorly at identifying the attributes in those classes; thus, they should not be considered as \textit{distinct}.

% We inspected the mean of participants' score and the standard error for each class, shown in Figure \ref{fig:surveyBarPlot}. 
% However, the mean score is not a true indicator of participants' performance, since different questions have different number of choices in them. An apt indicator of the performance is the difference between the mean score and expected score of random guess ($\mu_{r}$). The higher the mean score with respect to the \textit{expected} random score, the more likely the hypothesis \textbf{H0} should be rejected.

To substantiate our claims about \textit{distinct} attributes,
% To identify the attributes which are \textit{distinct} from a human's perspective, 
we ran a one-tailed \textit{t}-test for each question class, comparing participants' score with the expected score of the random guess. A \textit{significance level} ($\alpha$) of 0.05 was used, which corresponds to confidence interval of 95$\%$. As shown in Table \ref{tbl:pvalue}, attribute classes Q4, Q11 and Q12 have \textit{p}-values higher than the threshold of 0.05, and hence, cannot be regarded as \textit{distinct} attributes. In contrast, the 10 other classes have \textit{p}-values significantly less than 0.05, implying that our hypothesis is supported. Consequently, we consider the attributes belonging to these classes as \textit{distinct}.

% one-sided t
% \todohere{Table for p-value.}

\begin{table}
\centering
\caption{ \small Participant Scores and \textit{t}-Test Results for each Attribute-Class. $^{*}$ denotes \textit{distinct} attribute classes.}
\begin{tabular}{|l|p{0.9cm}|p{1.4cm}|p{1cm}|l|l|}
\hline
\textbf{Classes} & Sample Mean & \textit{Expected} Rand. Score & Std. Error & \textit{t}-Stat & \textit{p}-Value \\ \hline
 Q1$^{*}$ & 84.0 & 50.0 & 5.237 & 6.492 & $<$ 0.0001 \\ 
 Q2$^{*}$ & 74.0 & 33.33 & 6.266 & 6.492 & $<$ 0.0001 \\ 
 Q3$^{*}$ & 70.0 & 50.0 & 6.546 & 3.055 & 0.002 \\ 
 Q4 & 24.0 & 20.0 & 6.101 & 0.656 & 0.258 \\ 
 Q5$^{*}$ & 94.0 & 33.33 & 3.393 & 17.883 & $<$ 0.0001 \\ 
 Q6$^{*}$ & 78.0 & 20.0 & 5.918 & 9.80 & $<$ 0.0001 \\ 
 Q7$^{*}$ & 88.0 & 25.0 & 4.642 & 13.571 & $<$ 0.0001 \\ 
 Q8$^{*}$ & 68.0 & 20.0 & 6.664 & 7.203 & $<$ 0.0001 \\ 
 Q9$^{*}$ & 70.0 & 20.0 & 6.546 & 7.638 & $<$ 0.0001 \\ 
 Q10$^{*}$ & 60.0 & 25.0 & 6.999 & 5.001 & $<$ 0.0001 \\ 
 Q11 & 54.0 & 50.0 & 7.120 & 0.562 & 0.288 \\ 
 Q12 & 58.0 & 50.0 & 7.051 & 1.135 & 0.131 \\ 
 Q13$^{*}$ & 75.0 & 50.0 & 6.103 & 4.096 & $<$ 0.0001 \\ \hline
\end{tabular}
% \vspace{1cm}
Number of samples (\textit{n}) = 50 \hfill
\label{tbl:pvalue}
\vspace{-0.2in}
\end{table}

\subsection{Discussion}
\label{subsec:survey1Discuss}

\begin{figure}
\vspace{0.1in}
    \centering
    \includegraphics[trim= 20 0 150 0, clip, width=.48\textwidth]{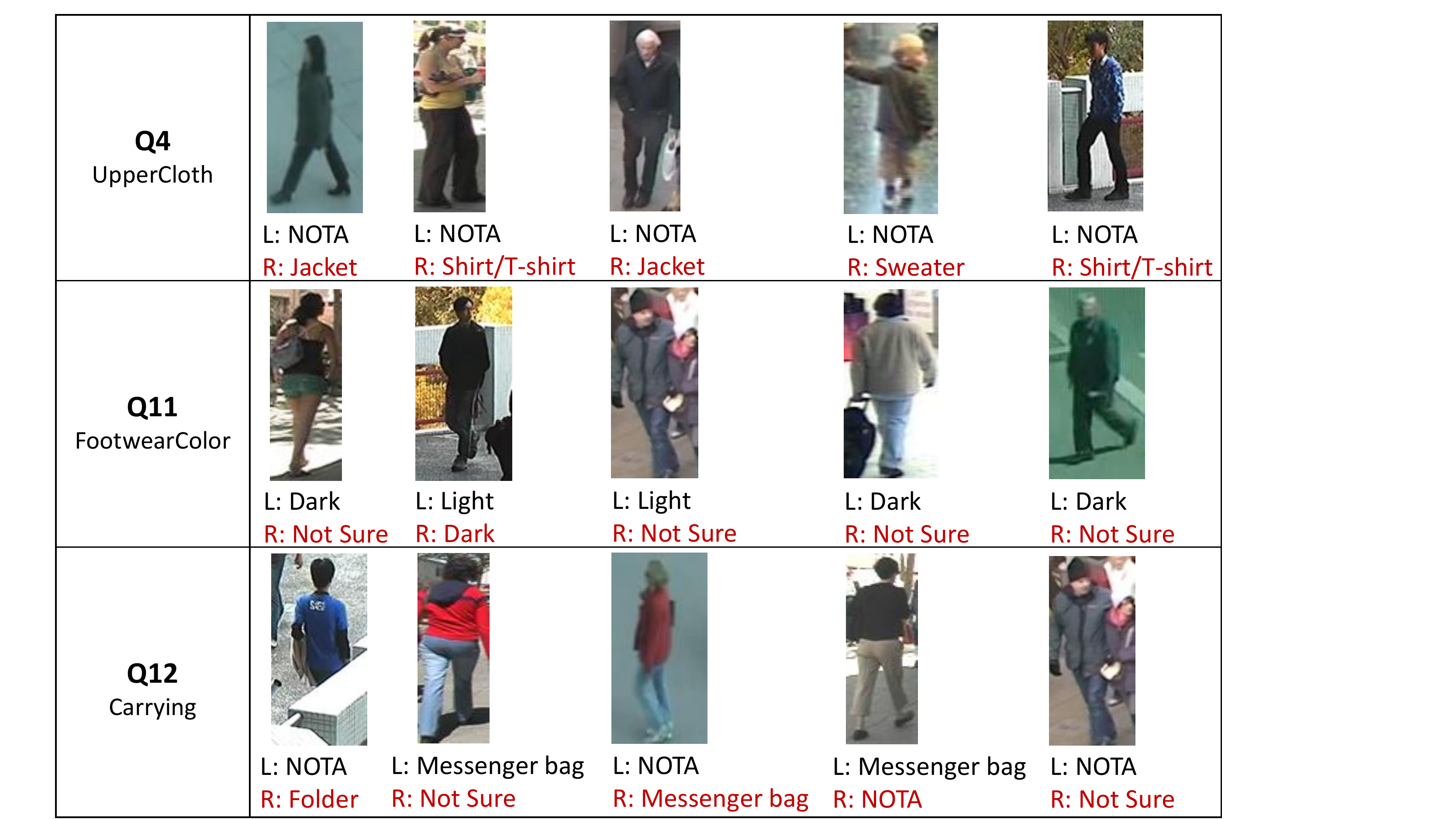}
    % \caption{\small Given a query image, the user describes the attributes of the person. The computer/robot retrieves the \textit{top-5} most relevant images from the gallery.}
    \caption{\small Randomly chosen \textit{incorrect} responses from the survey for Q4, Q11 and Q12 attribute class. `L' denotes label data and `R' denotes users' response. `NOTA' - None of the above.}
    \label{fig:mistakes}
    \vspace{-0.2in}
\end{figure}

% Based upon the \textit{t}-test study, we can conclude that human's are able to consistently predict the attributes belonging to class Q1, Q2, Q3, Q5, Q6, Q7, Q8, Q9, Q10 and Q13. This result qualifies them as \textit{distinct} attributes and should be used as input to the zero-shout re-ID system in real world applications.
% This result qualifies them as \textit{distinct} attributes and we apply them for zero-shot re-ID in our next section.
% \todohere{change the first line.}
% We believe there are two main reasons behind certain attributes not being recognized significantly better than random guess:
% We believe that there are two main reasons for the performance rift in case of certain attribute classes: 
We believe there are two main reasons behind humans performing only marginally better than random guess for certain attributes:
a) inconspicuous attributes and b) linguistic biases. First, some attributes are inherently hard to notice or remember. For example, in Figure \ref{fig:mistakes}, we can observe that most people are unsure about the `footwear-color'. 
% The second reason is more daunting. 
Second, different people can have differences in their choice of words for a certain visual attribute. This is referred to as linguistic bias \cite{echols2004identification}, and may arise from cultural, regional or gender differences. For instance, in Figure \ref{fig:mistakes}, for the first image Q4 class, the person seems to be wearing a `Jacket', however, the label suggests otherwise. Presence of linguistic biases in the labelled data could degrade the re-ID performance in real-world applications. 

\textbf{Generalization:} Since the survey was conducted with a few number of participants from a particular section of the society (graduate students at Cornell University), the conclusions regarding the \textit{distinct} attributes may vary across different population samples e.g. policemen may focus on very specific attributes for identifying a person.
% not be generalized to the entire world population. 
However, the important point that we want to emphasize upon is that the uncertainty in attribute-inference remains a challenge and depending upon the application, such a study is important to identify the key attributes that should be used for training the learning module.

% We observed that the participants tend to choose the option that is closest to the truth if 

% e.g. in many cases, it seemed like the person was wearing a ``jacket", however, as per the \textit{expert} data, ``None of the above" was the correct option.

% This motivates us to use them in the attribute vector for zero-shot re-ID in our next section. 
% Notice that, ignoring Q11 and Q12 which correspond to ``FootwearColor" and ``Carrying" classes respectively, seems like a logical choice because 
% The class Q4 corresponds to ``UpperCloth" class. 
% that are distinct and 
% In order do avoid this complication, we propose to 
% In this section, we describe our study to identify distinct attributes that should be used for 

\subsection{Limitations}
\label{subsec:limitation}

Although, our experimental setup is designed to simulate a realistic problem setting, there are certain limitations to this study which motivate future work. 
% After the survey, most participants reported that the resolution of the images was poor and their performance would improve with better quality images. 
After the survey, most participants reported that their performance was obstructed due to the low resolution of the images, as the datasets were mostly obtained from surveillance videotapes.
In an actual robotics application of zero-shot re-ID, the user would probably recall more detailed features about the person of interest, such as facial features, which were not used in our study. In this work, we were mainly limited by the attribute dataset. However, if a more descriptive dataset is available in the future, one could conduct the same study to identify the \textit{distinct} attributes and use them for training and testing ZSL models.

% \todohere{Ask Mark about this!}

% and use them in our KRR+DaRe model to enhance re-ID performance.

% This is mainly because we were limited by the attributes provided in PETA dataset and  
% In a real-world application, for example a search-and-rescue mission,  

% 1. Real world high resolution of eye.
% 2. In the study people remember the questions after a certain point.

% \todohere{Mark's comments included till here.}

\section{RE-ID WITH NON-EXPERT ATTRIBUTES}
\label{sec:humanRobotExperiment}
% \todohere{Maybe change title.}
% \todohere{Revise Section.}
% \todohere{Remove results for KRR from this section.}

% Having established the distinctiveness of certain attributes in Section \ref{sec:attributeStudy}, we utilize them for conducting a re-ID experiment, where the attributes are provided by a human user. As a first step, we train our re-ID model with \textit{distinct}, labelled attributes, available from the PETA dataset. In contrast with Section \ref{sec:zeroshotReid}, here we ask people to provide the attributes for the query images and match it to corresponding images from the gallery. This section describes the experiment design and the results.

% Having established the distinctiveness of certain attributes in Section \ref{sec:attributeStudy}, we conclude that not all the attributes are reliable in a real-world zero-shot re-ID applications. Therefore, it makes sense to only use the \textit{distinct} attributes for training and testing the ZSL models. 

Having established the distinctiveness of certain attributes in Section \ref{sec:attributeStudy}, we conclude that not all the attributes can be reliably observed by humans, assuming they do not have access to the query images. In the context of zero-shot re-ID, incorrect and missing attributes could potentially penalize the accuracy of ZSL models; thus, it is important to only rely upon the \textit{distinct} attributes. 
% To assess the performance of ZSL models in a realistic setting, we conducted a zero-shot re-ID experiment with human-annotated data. 
To justify this, we experiment with attributes provided by our human observers (denoted as non-expert attributes), in comparison to the expert-annotated ones, provided in PETA. 
This section describes the experimental design steps and results.

% Having established the distinctiveness of certain attributes in Section \ref{sec:attributeStudy}, we utilize them for conducting a re-ID experiment, where the attributes are provided by a human user. 
% This section describes the experimental design steps and the results.

% \subsection{Dataset}

% We have performed the human-robot experiment with VIPeR dataset \cite{gray2007evaluating}. The training and validation splits are same as Section \ref{subsec:reidPerformance}, while the test set consists of 50 pedestrian image pairs, randomly picked from the original test set. However, we only utilize the attributes belonging to class Q1, Q2, Q3, Q5, Q6, Q7, Q8, Q9, Q10 and Q13 from Table \ref{tbl:attributes} for training the KRR model.

% We have used the VIPeR dataset \cite{gray2007evaluating} for the human-robot experiment.

\subsection{Experiment Design}

% The proposed experimental setup is exactly as in a typical zero-shot re-ID setting, where based upon the attributes of the query images, we need to match them to the gallery.
The proposed experimental setup is exactly as in a typical zero-shot re-ID setting, where attributes are used to match the query images to the gallery.
The only difference here is that the attributes are now reported by humans. To fetch the attributes, we designed a survey, similar to the one in Section \ref{sec:attributeStudy}, where participants can label the attributes corresponding to the query images. The participants are allowed to examine each image for limited time (15 seconds), which is followed by the questionnaire about the attributes. 
However, unlike Section \ref{sec:attributeStudy}, the questionnaire only consists of the \textit{distinct} classes i.e. Q1, Q2, Q3, Q5, Q6, Q7, Q8, Q9, Q10, and Q13. Although, using only the \textit{distinct} classes does not guarantee no missing and incorrect attributes, we expect the use of \textit{distinct} attributes to be more reliable.
\subsection{Dataset and Implementation Details}
We performed the experiment using VIPeR dataset \cite{gray2007evaluating}. The training and validation splits are the same as Section \ref{subsec:reidDataset}, while the test set consists of 50 pedestrian image pairs, randomly selected from the original test set. 
The ZSL models are always trained with expert-annotations, obtained from PETA dataset. However, the testing is performed based on both expert and non-expert annotated attributes.
Non-expert annotations are retrieved from the survey, conducted with 10 participants, each responding to the questionnaire for 5 images; that covers all the 50 query images in the test set.
The participation for the survey was voluntary. All of them are graduate students, enrolled at Cornell University, belonging to the age group of 20-30 years. In addition, it was ensured that the participants of this survey are disjoint from the participants of the study in section \ref{sec:attributeStudy}.
% To evaluate the performance of ZSL models with user-input, we trained the models with \textit{distinct}, labelled attributes and deep-features obtained from DaRe network.
% We train ZSL models with labelled, \textit{distinct} attributes and deep-features produced by DaRe network, and evaluate the performance with user-input data.
% For evaluating the performance of ZSL models with user-input data, we train them based upon only the \textit{distinct} attributes. 
A few missing attributes can be encountered in the responses collected from participants; thus, during testing,
% To fill these missing fields in the attribute vector, we resort to mean-imputation, where the average of the training set is used for filling in.
the mean of training data is used for filling in missing attributes.

\subsection{Results}

% \todohere{Check this subsection with Harry.}

% % VIKRAM wrote the following part.
% We evaluate and compare the performance of ZSL models with labelled and user-input attributes\footnote[4]{The participant responses and the images used for the re-ID experiment can be viewed at \href{https://github.com/vikshree/humanRobotReid}{\texttt{https://github.com/vikshree/humanRobotReid}}}, and the results are reported in Table \ref{tbl:humanReidPerformance}.
% First, we evaluate 
We test the performance of ZSL models with: 1) Expert-annotated full attribute set, 2) Expert-annotated \textit{distinct} attributes, and 3) Non-Expert annotated \textit{distinct} attributes. The results are presented in Table \ref{tbl:humanReidPerformance}.

Comparing the performance in first and second cases, we observe that DAP and ESZSL achieve superior performance when they are trained and tested with the \textit{distinct} labelled attributes.
% compared with all 44 labelled attributes. 
EXEM achieves similar accuracy in both the cases. 
% This suggests that using less number of attributes leads to better performance.
This suggests that using less number of attributes does not hurt the performance and indeed improves it for certain models.
One explanation for this is that there is positive correlation between attributes being \textit{non-distinct} and having a lower ability to discriminate between different people. For example, footwear color is not a \textit{distinct} attribute, based on our survey in Section \ref{sec:attributeStudy}. Also, most people tend to wear dark colored footwear, thus diminishing the discriminative power of that attribute.
Therefore, removing `footwear-color' attribute will not degrade the re-ID performance. In the contrary, it could improve the performance by simplifying the learning task, since the size of one of the inputs to the model has reduced.

\begin{table}
\vspace{0.1in}
\centering
\caption{ \small \textit{Cumulative Matching Characteristic} accuracy for zero-shot re-ID on VIPeR dataset, consisting of 50 person-image pairs.}
% Superior results are shown in \textbf{bold}.
% \todohere{Search foe better hyperparameters for KRR model with 44 and 34 attributes. Train on validation set.} 
% }
\tabcolsep 3.5pt
\begin{tabular}{|l|l|l|l|l|l|}
\hline
~ & \textbf{Type of Test Data} & \textbf{Rank-1} & \textbf{Rank-5} & \textbf{Rank-10} & \textbf{Rank-25} \\ \hline
 
 \multirow{3}{*}{\rotatebox[origin=c]{90}{DAP}} & Expert, all attributes & 28.0 & 68.0 & 86.0 & 96.0\\
   & Expert, \textit{distinct} & 28.0 & 72.0 & 88.0 & 94.0\\
   & Non-Expert, \textit{distinct}& 22.0 & 44.0 & 72.0 & 88.0 \\
 \hline
 
   \multirow{3}{*}{\rotatebox[origin=c]{90}{ESZSL}} & Expert, all attributes & 28.0 & 66.0 & 90.0 & 100\\
   & Expert, \textit{distinct}& 30.0 & 76.0 & 90.0 & 100 \\
   & Non-Expert, \textit{distinct}& 22.0 & 56.0 & 80.0 & 100 \\
 \hline

   \multirow{3}{*}{\rotatebox[origin=c]{90}{EXEM}} & Expert, all attributes & 32.0 & 60.0 & 78.0 & 98.0\\
   & Expert, \textit{distinct}& 22.0 & 66.0 & 76.0 & 96.0\\
   & Non-Expert, \textit{distinct}&  22.0 & 60.0 & 76.0 & 96.0 \\
 \hline

%   \multirow{3}{*}{\rotatebox[origin=c]{90}{KRR}} & Label Att. (44) & 22.0 & 52.0 & 62.0 & 82.0\\
%   & Label Att. (34) & 18.0 & 56.0 & 68.0 & 86.0\\
%   & User Att. (34) &  16.0 & 56.0 & 68.0 & 92.0 \\
%  \hline

\end{tabular}
\label{tbl:humanReidPerformance}
\vspace{-0.1in}
\end{table}

The third case revealed the uncertainties that could arise during the deployment of the re-ID system for real-world applications. Comparing the participant responses\footnote[4]{The participant responses and the images used for the re-ID experiment can be viewed at \href{https://github.com/vikshree/humanRobotReid}{\texttt{https://github.com/vikshree/humanRobotReid}}} 
with expert-annotated, \textit{distinct} attribute data shows that on an average, there are 12 (out of 34) attributes that are either missing or incorrect. 
Consequently, we observe that the performance is noticeably diminished with non-expert annotations as compared to the expert-annotated case, for all three ZSL methods. 
In contrast to the current scenario, where the participants only annotate 34 \textit{distinct} attributes, using all 44 non-expert annotations in the ZSL models would have led to an even higher number of missing and incorrect values, thus impeding the performance even further.
% In contrast to the current scenario, using all 44 attributes in the third case would have led to an even higher number of missing and incorrect attributes, thus impeding the performance even further. 

% Also, the accuracy the KRR model in inferior to the other methods, however, the results with and without user-input are competitive, implying robustness. 

Moreover, we observe that ESZSL achieves an impressive Rank-25 accuracy of 100, even with non-expert annotated attributes. 
% This implies that the search space can be reduced to half of its original size, while still being confident that the person of interest is in the reduced search volume.
For a test set with 50 images, this implies that the search space can be shrinked to half of its original size, while still being confident that the POI is in the reduced search volume.

\section{CONCLUSION}
% \todohere{Revise Section}
% \todohere{It seems you developed a better architecture and did a cool experiment. So what?}

% % VIKRAM starts here
% \todohere{Harry: May be add some more future work!}

% In this paper, we take an attribute-based route for the zero-shot re-ID problem. 
In this paper, an attribute-based zero-shot re-ID solution is developed. 
By combining learned deep feature representation that renders high discrminative capability across identities, and representative ZSL models, we achieve state-of-the-art performance.
% We utilize a state-of-the-art feature representation that renders high discriminating capability among people of different identities. The approach achieves state-of-the-art performance with pre-existing ZSL models. 
Our human-subject study suggests that the task of estimating attributes encompasses uncertainties in the real-world. Consequently, we demonstrate that not all attributes are reliable for the re-ID task, in a realistic scenario. 
Finally, we evaluate practical sensitivities of the approach in realistic robotics applications by comparing ZSL methods tested with non-expert annotations vs expert-annotated attributes.  
% Finally, we evaluate the accuracy of different zero-shot re-ID approaches, tested with human-annotated attributes vs expert-annotated ones, since it exemplifies a more realistic, robotics application of the system. 
To the best of our knowledge, this is the first experimental evaluation of zero-shot re-ID methods where attributes are reported by participants, entirely on the basis of their recollection of the query images.
We found that the performance of some ZSL models can be severely impaired with incorrect and missing attributes in the wild. 
% Thus, it is necessary to carefully choose only the reliable attributes that humans can identify and recall consistently.
Thus, to execute zero-shot re-ID in real-world applications, it is necessary to take the uncertainty/unreliability in attribute annotations into account, for example, by only utilizing the \textit{distinct} attributes in our model. 

\bibliographystyle{IEEEtran} 
\bibliography{egbib} 

\end{document}